\pdfoutput=1

\documentclass[11pt]{article}

\usepackage[]{ACL2023}

\usepackage{times}
\usepackage{latexsym}
\usepackage{graphicx}
\usepackage{csquotes}
\usepackage{listings}
\usepackage{enumitem}

\usepackage[T1]{fontenc}

\usepackage[utf8]{inputenc}

\usepackage{microtype}

\usepackage{inconsolata}

\usepackage{soul}

\usepackage{cleveref}

\title{Debug Smarter, Not Harder: AI Agents for Error Resolution in Computational Notebooks}

\author{
\bf{Konstantin Grotov}$^{1, *}$, \bf{Artem Borzilov}$^{2}$, \\ \bf{Maksim Krivobok}$^{2}$, \bf{~Timofey Bryksin}$^{1}$, \bf{Yaroslav Zharov}$^{1}$ \\ 
\textsuperscript{1}JetBrains Research, \textsuperscript{2}JetBrains \\ \textsuperscript{*}\texttt{konstantin.grotov@jetbrains.com}
}

\usepackage{xcolor}

\definecolor{jbcolor}{HTML}{6B57FF}
\newcommand{\userstudyquestion}[2]{\textit{Q#1\textcolor{jbcolor}{``#2''}}}

\begin{document}
\maketitle
\begin{abstract}
Computational notebooks became indispensable tools for research-related development, offering unprecedented interactivity and flexibility in the development process. 
However, these benefits come at the cost of reproducibility and an increased potential for bugs.
With the rise of code-fluent Large Language Models empowered with agentic techniques, smart bug-fixing tools with a high level of autonomy have emerged.
However, ose tools are tuned for classical script programming and still struggle with non-linear computational notebooks.
In this paper, we present an AI agent designed specifically for error resolution in a computational notebook. We have developed an agentic system capable of exploring a notebook environment by interacting with it---similar to how a user would---and integrated the system into the JetBrains service for collaborative data science called Datalore.
We evaluate our approach against the pre-existing single-action solution by comparing costs and conducting a user study. Users rate the error resolution capabilities of the agentic system higher but experience difficulties with UI. We share the results of the study and consider them valuable for further improving user-agent collaboration. 
\end{abstract}

\section{Introduction}
Computational notebooks have become a popular medium for development during the last decade, 
especially for data analysis, machine learning~\cite{perkel2018jupyter}, and creating educational~\cite{barba2019teaching} or scientific content~\cite{perkel2021ten}.
One of the main features of computational notebooks is their statefulness---thus the notebook cannot be described only by its cells, but additionally, runtime information is required.
The statefulness allows to work iteratively with the runtime in an additive manner and thus to efficiently go through hypotheses~\cite{rule2018exploration}. However, it causes high code entanglement~\cite{ramasamy2023workflow, rule2018exploration} and, therefore, a higher number of errors in the code. As a result, notebooks are struggling with low reproducibility rates. After a re-run, they come to the same results with a 4\% probability~\cite{pimentel2019large}, and 75\% of them could not be executed without exceptions~\cite{pimentel2021understanding, pimentel2019large}. The resulting debugging distracts developers from the actual task.

Large Language Models (LLMs), such as GPT-4~\cite{OpenAI_GPT4_2023}, Mixtral~\cite{jiang2024mixtral}, or Code Llama~\cite{roziere2023code} recently demonstrated advanced capabilities in solving complex code-related problems, such as code generation~\cite{ni2023lever, wu2023autogen}, debugging~\cite{tian2024debugbench, bouzenia2024repairagent}, and issue solving~\cite{zhang2024autocoderover, yang2024swe}. However, there is a lack of studies applying such models to notebooks. The difficulty lies in the stateful nature of the notebook. Since the notebook requires runtime information to represent its exact current state, it is hard to gather the context for an LLM, as passing the entire runtime information is impossible due to the context size limitations.

AI agents allow LLMs to interact with such an environment iteratively. 
An agent can explore the environment and achieve the goal autonomously, enabling it to adjust its actions based on the received feedback.
Such agents have shown abilities to engage with software engineering tasks~\cite{wang2024executable, tufano2024autodev,si2024design2code, yang2024if}, interact with web environments~\cite{drouin2024workarena,zhou2023webarena}, or operate embodied agents~\cite{wang2023describe}. 

In this work, we present an AI Agent for error resolution in computational notebooks. The proposed agent was integrated into Datalore,\footnote{Datalore: \url{datalore.jetbrains.com}} a JetBrains product for collaborative data science that allows development in cloud-hosted computational notebooks. We design the agent to be capable of creating, editing, and executing cells. This approach utilizes the notebook's natural interactivity and allows gradual expansion of context. 

The main contributions of our paper are:
\begin{itemize}[noitemsep,nolistsep]
    \item An LLM-based AI Agent integrated into Datalore.
    \item A cost analysis of the proposed agent. 
    \item A user study on developers' experience with agentic systems in their workflows.
\end{itemize}

In~\Cref{sec:system-design}, we describe the overall design of our agentic system. After that, in~\Cref{sec:evaluation}, we evaluate our agent and discuss the results. Finally, we describe the limitations and conclude our work.

\section{System Design}
\label{sec:system-design}
In this section, we will delve into the proposed system's architecture. The system contains three parts: an agent, an environment, and a user interface. The agent is a stateful back-end service responsible for orchestrating the communication between the LLM and the notebook, storing prompts, and converting LLM predictions into actions in the environment. 
The environment is the computational notebook that---in addition to being fully functional---is responsible for executing actions provided by the agent and providing corresponding observations. The user interface defines how programmers interact with the system as a whole.

The goal of the system was to conduct the necessary code changes and cell executions to \textit{solve}\footnote{In formal terms, agents can solve errors by either commenting on or deleting code in a cell. However, for us, resolving errors means accurately identifying and resolving the underlying cause of the error.} the given runtime exception.

\subsection{AI Agent}
To set up an AI agent, it is necessary to choose an LLM, a memory stack storing the interaction history, a strategy for solving the particular problem, and a set of tools that will be available to the agent for interacting with the environment. 
The tools we chose are described in the~\Cref{subsection:tools}, as the environment provides them. In this subsection, we concentrate on the other parts of the agent.

As an LLM for our agent, we chose the \texttt{GPT-4-0613} foundation model with the ability of function calling. We selected this specific model based on its reliable performance in producing function calls as of April 2024. On each generation step, the LLM is prompted with the history of previous LLM generations, as well as observations from the environment. This constitutes the memory stack.

The strategy consists of the system prompt for the LLM and the algorithm for converting an LLM prediction into a tool call. 
The system prompt, in our case, contains the description of the general goal (which is error resolution), the tools, and the guidelines. As the guidelines, we described the hacks prohibited during the workflow and encouraged the agent to explore the environment and to avoid actions with large outputs. We considered an action a hack if it technically ablates errors instead of resolving them. For example, deleting the code cell that caused an error is a hack.

We used reflection, akin to~\citet{shinn2024reflexion}, as the algorithm for choosing the next action.
In this algorithm, at each step, the LLM is prompted to reflect on the outcomes of the previous actions before selecting the next tool to call.

The AI Agent was developed as a service that communicates with the environment using HTTP requests. After the environment makes an initial request, the service creates a new stateful instance of the agent. Once the agent generates the next step, it is translated into a tool-calling instruction and sent back as a response. If the proposed tool is not ``Finish'', the agent waits for another request from the environment with the new observation. Otherwise, the agent process is terminated, and the previous session is no longer available. The agent follows the strategy until it solves the error, reaches the maximum number of iterations (15), or exceeds the maximum response timeout of 15 minutes.

\subsection{Tools and Environment}
\label{subsection:tools}
During error resolution, the agent collects new observations from the environment using various tools. A tool is an action available for the agent. Then, on the environment side, the particular tool call is processed, and the result is returned for the agent to adapt and continue the strategy loop.

The environment in this context is a computational notebook (similar to Jupyter notebooks), which provides an interactive interface for writing and executing code in the cells. The environment supports Python and offers features like inline plotting, markdown support, and the ability to execute shell commands, making it a versatile tool for data analysis and development.

We extended the environment with tools to allow the agent to interact with the notebook environment in a manner natural to developers, seamlessly integrating it further into the workflow. The proposed list of tools includes the following: creating, editing, and executing cells. Additionally, the ``Finish'' action was introduced, enabling the agent to stop independently. This action allows the agent to halt its activities before reaching the maximum iteration count. With these tools, the agent can explore the environment even beyond the current notebook state. For example, the agent can execute the \texttt{!ls} code cell to explore files outside the notebook.

\begin{figure}[t]
\centering
\includegraphics[width=7cm]{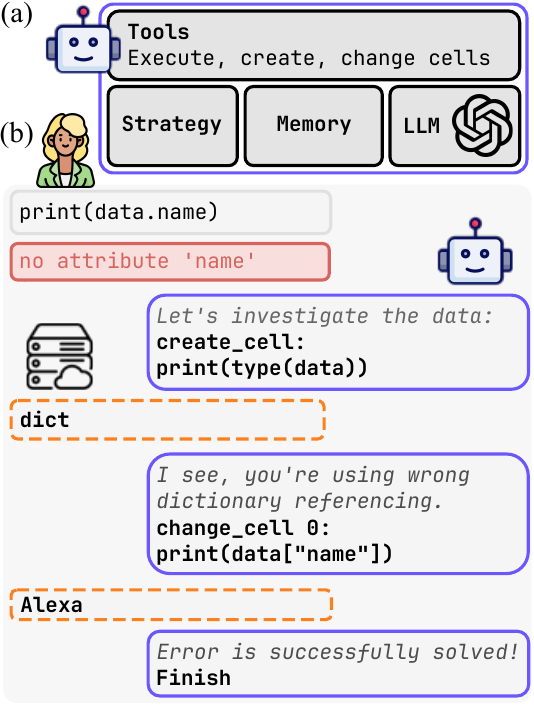}
\caption{(a) The components of AI agent. (b) Interactions of the AI Agent during error resolution. Once an exception appears, the agent starts to interact with the notebook environment to get valuable context and resolve the error.}
\vspace{-0.1cm}
\label{fig:ai-agent-scheme}
\end{figure}
\begin{figure*}[t]
\centering
\includegraphics[width=1\textwidth]{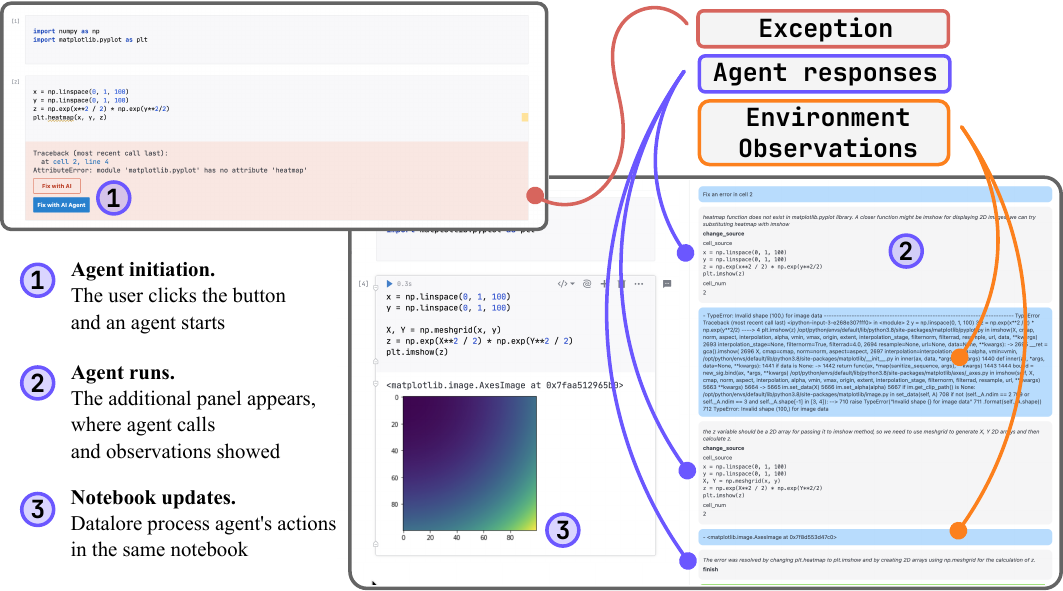}
\caption{AI Agent in the Datalore notebook. Once an error appears, the user can initiate the work of an agent, and it will iteratively resolve the error and reflect on its actions respectively.}

\label{fig:screenshot-system}
\end{figure*}

The environment initiates the agent's workflow by sending the error stack trace with the corresponding cell number and the notebook cells source without outputs. After receiving the response from the agent, the proposed tool is executed and responded to with the cell output as the observation. The schematic diagram of the automatic error-solving workflow is shown in Figure~\ref{fig:ai-agent-scheme}. All prompts can be found in~\Cref{sec:appendix-prompts}, and tool descriptions are in the supplementary materials.

\subsection{User Interface}
We incorporated the user-agent interaction in the computational notebooks available in Datalore. Once an error occurs in a cell, the additional ``Fix with AI Agent'' button appears, which allows one to initiate the error resolution process. After the user clicks on this button, an additional panel appears on the right side of the screen, displaying the chat between the agent and the environment. Every action the agent proposes is displayed in the chat with an additional explanation by the agent of why it chose it. Simultaneously with the changes in the chat, the actions are executed in the notebook environment, and cell outputs are sent back to the agent as observations. The interface of the system is elucidated in Figure~\ref{fig:screenshot-system}.

\begin{figure*}[t]
\centering
\includegraphics[width=1\textwidth]{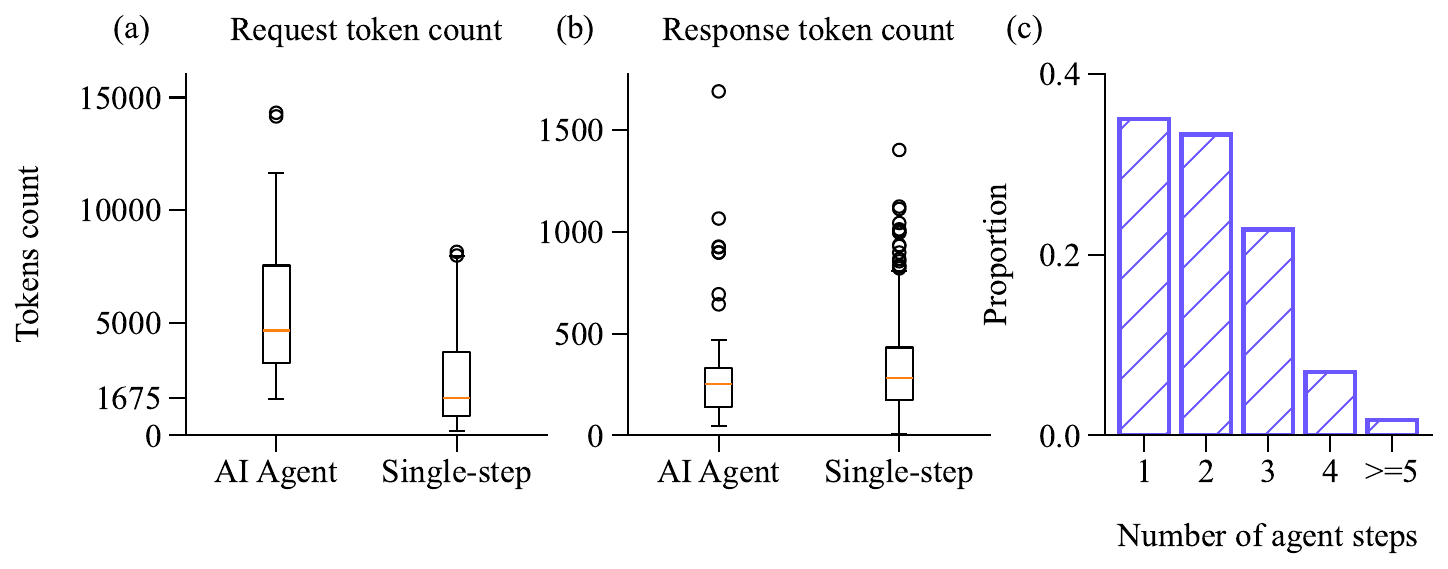}
\caption{AI Agent evaluation. (a), (b) Comparison of AI Agent token consumption with the single-action solution. (c) Distribution of steps needed for an agent to solve the error.}
\label{fig:costs-analysis}
\end{figure*}

\section{Evaluation}
\label{sec:evaluation}
We evaluated our system from two perspectives: system performance and user experience. For the former, we compared the costs of employing such an AI agent and the single-action solution already implemented in Datalore. For the latter, we conducted a user study to estimate the effect on the developers' subjective productivity and satisfaction with error resolution capabilities. 

\subsection{Cost Analysis}
For cost analysis, we compared our developed AI agent with the single-action solution. A single-action solution has already been implemented in Datalore as an LLM-powered feature for Python exception resolution. It was implemented using a similar user interface but without multiple iterations. The system uses chain-of-thought reasoning~\cite{wei2022chain} to identify the cause of the problem and generates the code to resolve the issue in the current code cell. As the input context of the single-action solution, Datalore uses the notebook code and the cell number where the error appeared. We calculated the costs of a single-action solution using real user statistics gathered from Datalore. The data contained the consumption of both request and response tokens after each error resolution. 

For the evaluation of the AI agent, we used a dataset of fine-grained Jupyter Notebook execution logs.\footnote{The dataset is currently unavailable since it is part of another paper under review, and more specific information will be shared afterward. In the meantime, the dataset can be accessed upon request.} The dataset included over 100 hours of logs, capturing all cell additions, executions, and deletions made when solving data science tasks in a hackathon. A total of 20 people participated in the experiment. The key feature of the dataset is that the developer's workflow in the notebook can be reproduced, which was very useful for our analysis. We utilized the dataset to reproduce the notebook errors and then resolved them using the AI agent. This allows us to evaluate our system on real errors that occurred during development and fine-tune our agent strategy to solve errors better.

When evaluating the agentic workflow, we considered the error successfully resolved if the cell executes without an exception after the agent is finished. We also manually checked error resolution logs to ensure that the agent did not use prohibited hacks. There were no such cases.
We logged the history of notebook-agent interaction during error resolution, based on which we calculated the costs. The tokens were divided into request and response, since their prices differ significantly.

Figure~\ref{fig:costs-analysis} (a-b), shows the consumption of request and response tokens, respectively. We observe that the agentic system consumes almost three times more input tokens and almost the same amount of response tokens compared to the single-action solution. This is due to the growing memory stack. While the consumption of input tokens is significantly higher for the agentic system, it is still acceptable for industrial use due to the cheapness of these tokens compared with the output ones. The average cost of the single error resolution for the AI agent is $\$0.22$, and for the single-action solution --- $\$0.09$. To mitigate the growing context and further decrease the cost difference, one can turn to the context caching techniques~\cite{monteiro2024xc}.

Panel (c) highlights the distribution of iterations the agent needs for error resolution. We discovered that most frequently the errors were successfully solved in just one step. Furthermore, we observed that the frequency of step sequences sharply declines after the third step, with very few cases requiring more than five.

\begin{figure*}[t]
\centering
\includegraphics[width=1\textwidth]{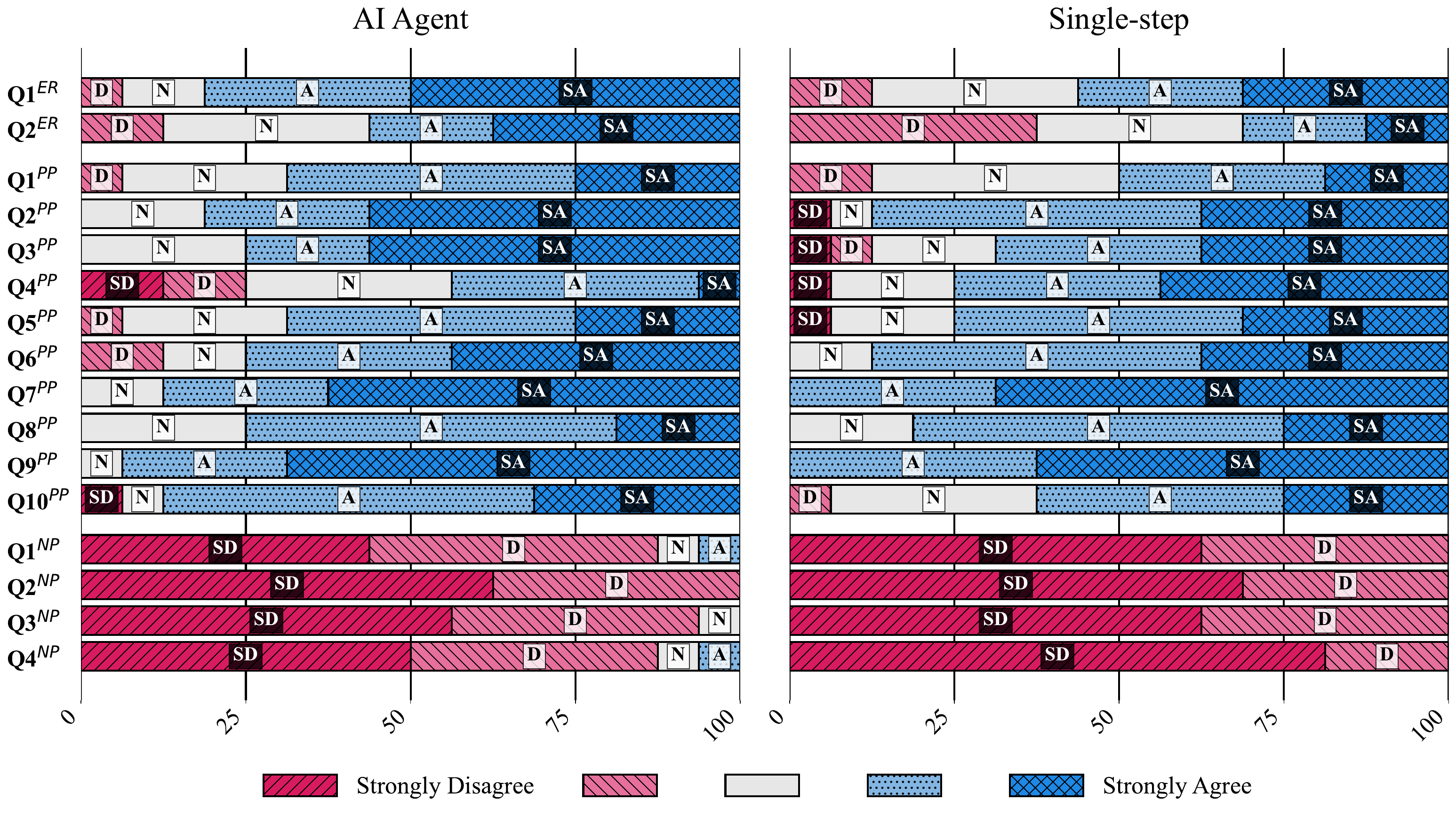}
\caption{The Likert diagram showed a comparison of question scores between the AI agent group and the single-step group. The questions are divided into three sections. The users rate the AI Agent error resolution capabilities higher, while the user experience worse.}
\label{fig:user-study}
\end{figure*}

\subsection{User Study}
To evaluate the effect of our agentic system on the developer workflow, we designed and conducted a user study. During the study, we measured the developers' subjective productivity and assessment of the systems' error resolution capabilities.
The study design included two groups of participants: one employing a single-action AI assistant and the other one using the AI Agent. We recruited participants within JetBrains without mentioning the group to which they were referred. As a result, we collected a sample containing 16 people in each of the two groups.

We offered both groups a data-filtering task designed to be completed within 30 to 45 minutes in Datalore. The task was to read the unstructured textual data, which had various mistakes that caused errors during pre-processing. The task could be solved using the Pandas Python package and the Python Standard Library. The full task description can be found in~\Cref{subsec:appendix-task-description}. Participants solved the task without supervision. They were allowed to solve the task at the time of their choice. However, we asked them to stop after 45 minutes.

After completing the task, each participant was asked to fill out a survey for qualitative analysis. In the survey, we asked questions divided into three categories: the system's error resolution capability (\textit{ER}) and both positive (\textit{PP}) and negative (\textit{NP}) productivity experiences while using the system. The System Usability Scale (SUS)~\cite{lewis2018system, brooke1996sus}, consisting of a 5-item Likert scale questionnaire (with items ranging from ``Strongly disagree'' to ``Strongly agree''), was used for qualitative analysis of user experience.

The results of the user study for each group of questions are shown in Figure~\ref{fig:user-study}.
We further discuss each group of questions separately.

\textbf{Error resolution capability.} 
The first group of questions is shown in the chart as \textit{ER} questions. The questions were: \userstudyquestion{1}{The system has a good understanding of errors} and \userstudyquestion{2}{The system effectively resolves errors as expected}. We note that the developers demonstrate a more positive perception of the agentic system and consider it more capable in terms of solving errors (mean value is $4.03 \pm 0.31$ for the AI agent and $3.41 \pm 0.49$ for the single-step system).

\textbf{Positive productivity feedback.}
The next set of questions labeled in the chart as \textit{PP} evaluated the positive subjective productivity feedback while using the system. The list of questions in this set is the following: \userstudyquestion{1}{I would let the system operate on my daily code and data}, \userstudyquestion{2}{I spend less time searching for information}, \userstudyquestion{3}{I complete my tasks more quickly}, \userstudyquestion{4}{I complete my tasks with less mental effort}, \userstudyquestion{5}{I have more time to engage in more interesting work-related tasks}, \userstudyquestion{6}{I think that I would like to use this system frequently}, \userstudyquestion{7}{I thought the system was easy to use}, \userstudyquestion{8}{I found the various functions in this system were well integrated}, \userstudyquestion{9}{I would imagine that most people would learn to use this system very quickly}, \userstudyquestion{10}{I felt very confident using the system}. We see that, on average, people highly rated both systems. For that, we calculated the average score using all questions in the group. We've got the average score in the group of $4.08 \pm 0.43$ for the AI agent and $4.10 \pm 0.35$ for the single-step one.
However, looking at the individual questions \textit{Q4} and \textit{Q10}, we note that while people rely on the AI agent more than on the simpler solution, mentally, it is harder to interact with the agent.

\textbf{Negative productivity feedback.}
The last group of questions labeled as \textit{NP} elucidates problems and difficulties experienced while using the system. For these questions, a higher score is worse. Here is the list of the questions: \userstudyquestion{1}{I found the system unnecessarily complex}, \userstudyquestion{2}{I think that I would need the support of a technical person to be able to use this system}, \userstudyquestion{3}{I found the system very cumbersome to use}, \userstudyquestion{4}{I needed to learn a lot of things before I could get going with this system}. We note that, on average, people rate the agentic system worse than the single-action one (mean value is $1.57 \pm 0.18$ for the AI agent and $1.31 \pm 0.09$ for the single-step one). We attribute this to an overloaded UI and overall new user experience of interacting with the system with a higher level of autonomy. We considered it this way because user interface and control were mentioned by participants in open questions. Some of them can be found in~\Cref{subsec:appendix-open-feedback}.

We calculated the SUS score~\cite{brooke2013sus} based on the users' answers. We note that both systems rated between ``good'' and ``excellent'' ($72.7$ for the AI agent versus $72.8$ for the single-step solution). The user study highlighted the strength of the proposed system---its ability to effectively resolve errors in computational notebooks, thus enhancing productivity during the data science workflow. However, the weakness of the proposed system lies in its user interface, which lacks user control and is difficult to understand.

\section{Conclusion and Future work}
\label{sec:conclusions-and-future-work}
In the present work, we have presented an agentic system for error resolution in computational notebooks. Our solution was integrated into JetBrains Datalore. The cost of running the system tripled, yet the cost stayed within the reasonable price range. The user study revealed many directions for further user-agent interaction research, such as ensuring the user's control over the agent or better visualization of the agent's actions. 

Utilizing smaller and cheaper models and more intelligent information retrieval holds potential for cost-efficient next generations of such systems. The context caching techniques also look promising in iterative agentic applications. To benefit the community, we publish used prompts and the answers from the user study.

\section{Limitations}
After the user study, we got many comments on improving the UI. The users mentioned that the agent took too much control of their workflow. While it performed actions with the appropriate reasoning, it was tough to keep track of them due to the speed of the agent's work. While it is a limitation of our system, which will be investigated more carefully, we found the general lesson of keeping the user in control useful for the community. Even though people generally agree to use the system for their own working tasks, we have not developed a secure sandbox. It is crucial to ensure the safety of their data and code while an agent explores the environment. 

Although the system showed higher costs than the single-step solution, the agent successfully found the solution in most cases within the first or second steps. Therefore, the agentic approach can be used to determine the valuable context for task-solving purposes, which can subsequently be incorporated into a single-step solution. 

Distinguishing between actual problem resolution and hallucination remains challenging algorithmically. Although the agent demonstrates effective error resolution in most observed cases, a quantitative evaluation of accuracy was not conducted. This presents a potential limitation, as the system may occasionally produce a seemingly correct solution that does not address the root cause of the error. Further research is needed to develop metrics that can automatically assess the correctness and relevance of the agent's solutions.

\bibliography{anthology,custom}
\bibliographystyle{acl_natbib}

\appendix

\section{Prompts}
\label{sec:appendix-prompts}
\subsection{System Prompt}

\begin{lstlisting}[frame=single, basicstyle=\ttfamily\scriptsize, xleftmargin=0pt, numbers=none]
You are a coding assistant which should help to solve user's error in computational notebook.
You should use functions to help handle the real time user queries and return outputs ONLY in the form of a valid JSON.

Remember:
  1. Keep trying for at least 10 steps before you stop. But if you think you solved the problem, you can finish right away.
  2. Use Python code only. When you need to explain what you did, write it as a comment in the code or in the `comment` field of the JSON.
  3. If you can fix the error without changing any code, do that. Don't edit the existing code or add new code unless you really need to.
  4. Use only the functions given to you. If you have many functions to choose from, pick the one that solves the problem quickest.
  5. Don't run the cell that caused the error. If you think you've fixed the error, run the "finish" function instead.
  6. If nothing shows up after you run a cell, that means there were no errors or outputs.

After you've done actions that you think have fixed the problem, run "finish" to say you're done. 
It's better to run a cell as is to fix errors than to change the cell's code.

You have a few ways of interacting with the environment:
  1. You can suggest new code for the existing cells, run it, and give the output.
  2. You can make a new cell with your own code, run it, and give the output.
  3. You can run any cell as is and give the output.
  4. If you're sure the error won't show up in the cell it was found in, you can run "finish".
\end{lstlisting}

\subsection{Initial Prompt Template}
\begin{lstlisting}[frame=single, basicstyle=\ttfamily\scriptsize, xleftmargin=0pt, numbers=none]
Here's a Jupyter notebook. It uses `{separator}` as a separator between cells. Note that cells indexes START FROM 1!
```
{notebook}
```
Error occurred in cell with num {cell_num}.
The error trace is the following:
```
{error}
```
Please resolve the error.
You must use only defined functions for solving the error. Return output only as a valid JSON.
YOU MUST NOT WRITE ANY COMMENTS / THOUGHTS / PLANNING OUTSIDE OF the "comment" JSON FIELD!
After you perform actions which should solve the error, use function finish to indicate that.
IF IT'S POSSIBLE TO SOLVE ERROR WITHOUT CHANGING THE CODE YOU MUST DO THAT!
IF YOU NEED ANY EXTRA INFORMATION GET IT VIA EXECUTION OF NEW CELL (CREATE IT, CHANGE SOURCE AND EXECUTE)
IF YOU WANT TO WRITE ANY COMMENT USE "comment" FIELD IN FUNCTION CALL AND NOWHERE ELSE!
YOU MUST NOT CHANGE FILES OUTSIDE OF THE NOTEBOOK BUT CAN EXPLORE THE ENVIRONMENT VIA EXECUTING NOTEBOOK CELLS.
Just adding try-except is not a solution. Commenting the code that produced error is not the solution.  You should propose only meaningful final solutions.
While exploring you must avoid large outputs, so be careful with prints.
\end{lstlisting}

\section{User Study Artifacts}
\label{sec:appendix-user-study}

\subsection{Data Filtering Task}
\label{subsec:appendix-task-description}

Several services simultaneously launched an AI assistant and agreed to jointly collect and analyze user feedback. Despite using the same LLM, the integration of feedback data faced challenges due to differences in data formats. Additionally, an issue emerged where timestamps were not logged correctly.

To facilitate the analysis, extract the user feedback data from the \texttt{aggregated\_logs.log} file located at the project root. This file contains merged logs from all participating services, structured with timestamp data preceding the JSON-formatted log entries.

The task is to create a \texttt{DataFrame} with the following structure:
\begin{itemize}[noitemsep,nolistsep]
    \item \texttt{hash}: \texttt{str}
    \item \texttt{service\_id}:  \texttt{int}
    \item \texttt{time}: \texttt{datetime}
    \item \texttt{is\_positive\_feedback}: \texttt{bool}
\end{itemize}

Further, analyze instances where timestamps are incorrectly logged (logged as `unknown` instead of the actual date) to identify potential patterns or systematic errors causing this issue. This might involve reviewing the formatting or encoding discrepancies among different service logs.

\textit{Please note that if you find yourself taking longer than 45 minutes, you should stop solving the task.}

\subsection{Open Feedback Responses}
\label{subsec:appendix-open-feedback}

Here are the selected answers for the following question: \textbf{Please share any comments or suggestions you have regarding aspects you disliked about the system or areas where you think the system can be improved.}

\begin{itemize}
    \item It's not always obvious which cell was edited by agent. like i tried to follow along with agent execution in an agent interaction window, but the texts fly quite fast, and once it's finished, you have to spend some time processing either the texts or your notebook to understand what actually happened.
    \item Overall, a problem I had with the AI, including the Compose or Code with AI, was that it overwrote the content of the entire cell. It would have been useful if I could have somehow specified to only edit within a selection to avoid unwanted changes further up in the cell. This could of course lead to the error not being resolved, but it could also serve as a way to ground the AI to the target task? Similarly to how AI in IDEs does code completion.
    \item Perhaps it would be beneficial to indicate more explicitly, what cell the agent is going to execute (in the user interface), and maybe cleanup the cells it created to launch its own code (mine created a cell in the end of notebook with "print(logs[:5])" or smth like this, and it stayed after agent's finish)
    \item It would also be great if there was a separate window where I could enter my request to the agent, not just being able to use it only in case of Errors
    \item A very obvious commentary, but it's slow. That's not a problem if you can work while it's thinking. The problem with that is that it's jumpy when everything is changing around you. It seems like there is no "protection" even for a cell you are now working on.
    \item A lot of time, I felt like I needed help without an explicit red error. It just wasn't doing what I wanted. I am not sure what UX is needed, or how it is possible to communicate desire to agents, but that would be very cool.

\end{itemize}

\end{document}